\pdfoutput=1
\documentclass[twocolumn]{article}

\usepackage{epsfig}
\usepackage{subcaption}
\usepackage{calc}
\usepackage{amssymb}
\usepackage{amstext}
\usepackage{amsmath}
\usepackage{amsthm}
\usepackage{multicol}
\usepackage{pslatex}
\usepackage{lmodern,charter}
\graphicspath{{./figs/}}
\usepackage{siunitx}
\usepackage{algorithm}
\usepackage{algpseudocode}
\usepackage{amsfonts}
\usepackage[capitalise]{cleveref}
\usepackage{pdfpages}

\begin{document}

\title{Parallel Parking: Optimal Entry and Minimum Slot Dimensions}

\author{
Jiri Vlasak$^1$$^2$\footnote{https://orcid.org/0000-0002-6618-8152}
, Michal Sojka$^2$\footnote{https://orcid.org/0000-0002-8738-075X}
, and Zden\v{e}k Hanz\'{a}lek$^2$\footnote{https://orcid.org/0000-0002-8135-1296}}
\date{{\small
$^1$Faculty of Electrical Engineering, Czech Technical University in Prague
\\
$^2$Czech Institute of Informatics, Robotics and Cybernetics, Czech Technical University in Prague
\\
\{jiri.vlasak.2, michal.sojka, zdenek.hanzalek\}@cvut.cz
}}

\null
\includepdf[pages=1,fitpaper,noautoscale]{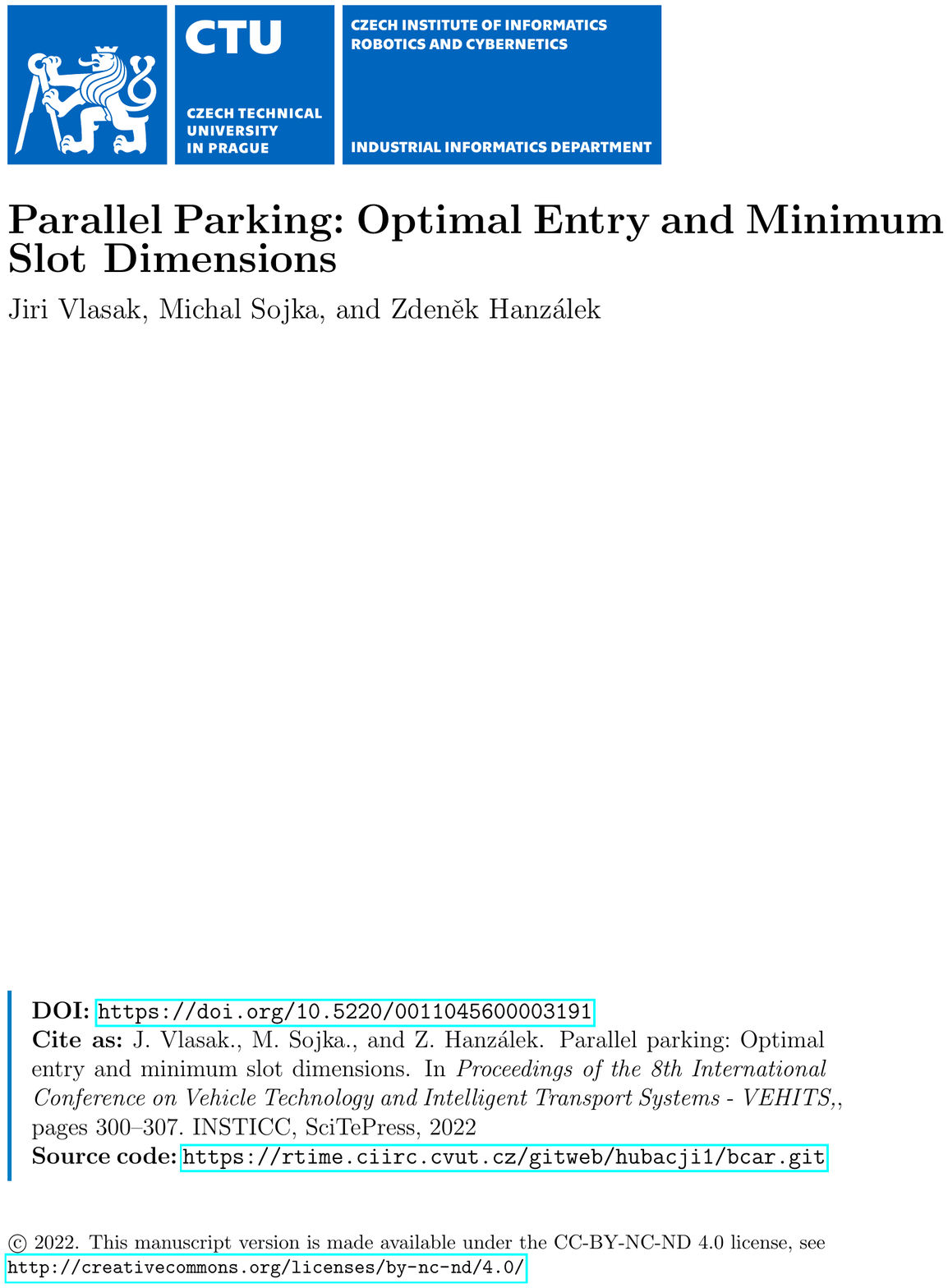}

\maketitle \normalsize \vfill

\abstract{The problem of path planning for automated parking is usually presented as finding a collision-free path from initial to goal positions, where three out of four parking slot edges represent obstacles. We rethink the path planning problem for parallel parking by decomposing it into two independent parts. The topic of this paper is finding optimal parking slot entry positions. Path planning from initial to entry position is out of scope here. We show the relation between entry positions, parking slot dimensions, and the number of backward-forward direction changes. This information can be used as an input to optimize other parts of the automated parking process.
\\\\
{\bf Keywords:} Automated parking, parallel parking}

\newcommand\ce[0]{C_{\text{E}}}
\newcommand\cep[0]{C_{\text{E}}^\text{P}}
\newcommand\ces[0]{\mathbf{\ce}}
\newcommand\ceas[0]{\mathbf{C_{\text{E}}^\text{A}}}
\newcommand\ceps[0]{\mathbf{\cep}}
\newcommand\cp[0]{C_\text{G}}

\section{\uppercase{Introduction}}
\label{s:intro}

Driver assistance systems experience a great expansion. Proposals for automated driving are topics of many teams of engineers and academics. One of the basic building blocks of automated driving is the (automated) parking assistant.

A common approach to the problem of path planning for parallel parking is to geometrically compute a path from the current (initial) position to the goal position inside the parking slot. When an optimization-based approach is used, the time or path length between initial and goal positions is usually minimized. But what if the given goal position is not chosen optimally?

In this paper, we reformulate the path planning problem of automated parking for a parallel parking slot into the problem of finding optimal entry and goal positions. We solve this problem for the given parking slot and car dimensions using a simulation approach. A feasible path is computed by simulating the car movement between these positions.

We show how we find the entry positions, the dependency between entry positions, parking slot dimensions, and the number of backward-forward direction changes, and how to find the minimum dimensions of a parking slot for the given car. The results of our simulations can be used as parameters in parking assistant algorithms to quickly decide about the possibility of parking and simplify the navigation toward the parking slot.

The rest of the paper is structured as follows. We define the parallel parking problem in \cref{s:p}. In \cref{s:isp}, we describe how we find the entry positions, how entry positions depend on direction changes, and how we compute the minimum dimensions of a parallel parking slot. In \cref{s:eval}, we perform computational experiments and compare our results to related works. We conclude the paper in \cref{s:concl}. The source code of our algorithm is publicly available as free and open source software\footnote{https://rtime.ciirc.cvut.cz/gitweb/hubacji1/bcar.git}.

\subsection{Related Works}

\cite{blackburn_geometry_2009} explains how to compute the minimum length of a parallel parking slot, to which the car can park in only one maneuver. In this paper, we consider parallel parking slots that could be shorter and therefore we allow multiple backward-forward direction changes.

\cite{zips_fast_2013} introduce two-phased constraint optimization algorithm to plan the parking maneuver for parallel, perpendicular, and angle parking slot. During the first phase, the algorithm computes a path from the goal position to a phase switching point, from which the car can leave the slot. In the second phase, the algorithm finds a path from the switching point to the initial position. In their follow-up work \cite{zips_optimisation_2016}, they present improvements to the algorithm for a path planning outside the parking slot. Our paper differs in the problem definition. We consider the parallel parking slot -- not the goal position -- as the input. Our approach guarantees the optimality of entry positions as opposed to the phase switching point.

\cite{vorobieva_automatic_2015} present three algorithms for geometric parking: (1) The algorithm for \emph{parking in one maneuver} computes path from the goal position by simulating forward movement with the maximum steering. This algorithm works only for long enough parking slots. (2) Parking in \emph{several parallel trials} algorithm starts by computing the path from initial position to some position partially inside the parking slot with the constraint of the car heading being parallel to the parking slot heading. Then, the algorithm continues by simulating forward-backward moves toward the parking slot with the constraint of the car heading being parallel to the parking slot heading when the car stops for the direction change. (3) Parking algorithm called \emph{several reversed trials} generalize the algorithm for parking in one maneuver. It computes the path in the reverse order to the parking process: the algorithm starts from the goal position and computes the path by simulating forward-backward moves with the maximum steering until the car leaves the parking slot.

\cite{li_implementation_2016} propose complete system for automated parking. Their planning algorithm computes a path for a parallel parking slot from three circle segments. The initial position is considered a part of the input to the algorithm and only one (backward) move is allowed.

The works above describe a geometric planner to find a path between initial and goal position. In this paper, we also use the geometric approach. However, we consider entry and goal positions as the result, which allows us to guarantee that the car can park from entry positions into the parking slot with the minimum number of backward-forward direction changes. Path planning from initial to entry position is out of scope this paper.

\cite{li_time-optimal_2016} present dynamic optimization framework for computing a time-optimal maneuvers for parallel parking. They compute trajectory from initial position to any parked position, i.e., a position of a car when the car is completely inside the parking slot. Compared to our approach, we do not consider the time. However, we compute a set of entry positions and return it as the result.

To solve the path planning problem of parallel parking for automated vehicle, \cite{jing_multi-objective_2018} introduce nonlinear programming optimization that minimizes multiple objectives such as path length, distance from the car front to the parking slot front, and the distance from the car center to the parking slot center. The initial position of the car is fixed and only one (backward) move is allowed. In this paper, we allow multiple backward-forward direction changes, but we minimize their count.

\section{\uppercase{Parallel Parking Problem}}
\label{s:p}

We define the used terminology and the parallel parking problem in this section.

\subsection{Definitions}

\emph{Car position} is a tuple $C=(x, y, \theta, s, \phi)$, where $x$ and $y$ are cartesian coordinates of the rear axle center, $\theta$ is car heading, $s\in\left\{-1,+1\right\}$ is direction of the movement (backward and forward respectively), and $\phi\in\left[-\phi_{\max},+\phi_{\max}\right]$ is car steering angle. Car positions are subject to a discrete kinematic model \cite{kuwata_real-time_2009} $C_{k+1} = f(C_k)$, $k\in \mathbb{N}$ where function $f$ is given by~\cref{e:1}:

\begin{equation}\label{e:1}
\begin{split}
       x_{k+1}&=x_k+ s_k\cdot \Delta\cdot cos(\theta_k)\\
       y_{k+1}&=y_k+ s_k\cdot \Delta\cdot sin(\theta_k)\\
       \theta_{k+1}&=\theta_k + \frac{s_k\cdot \Delta}{b}\cdot tan(\phi_k),
\end{split}
\end{equation}

where $\Delta \in \mathbb{R^+}$ is a positive constant we call the \emph{step distance}.

The \emph{direction change} is a change of the car movement direction $s$. We denote $\Gamma$ the number of direction changes.

\emph{Car dimensions} is a tuple $D=(w, d_\text{f}, d_\text{r}, b, \phi_{\max})$, where $w$ is the width of the car, $d_\text{f}$ and $d_\text{r}$ is the distance from the rear axle center to the front, respective back of the car, $b$ is the wheelbase (distance between the front and rear axle), and $\phi_{\max}$ is the maximum steering angle.

\emph{Car frame} $\mathcal{F}(C)$ is a rectangle given by car position $C$ and car dimensions $D$.

\emph{Minimum turning radius} is the radius $r$ of the circle traced by the rear axle center when the car moves with the maximum steering. It holds that $r=\frac{b}{\tan \phi_{\max}}$.

\emph{Curb-to-curb distance} is the diameter $d$ of the circle traced by the outer front wheel when the car moves with the maximum steering. It holds that $d=2\cdot\sqrt{\left(r + \frac{w}{2}\right)^2 + b^2}$.

\emph{Parallel parking slot} is a rectangle whose one side (that we call \emph{entry side}) is adjacent to the road. \emph{Parallel parking slot} is defined as a tuple $P = (p, \delta, W, L)$, where point $p\in \mathbb{R}^2$ is one corner of the rectangle on the entry side, $\delta$ is the direction vector of the entry side relative to $p$, $W > w$ is parking slot width greater than the width of the car, and the length of the parking slot, equal to the length of the entry side, is $L\ge d_\text{f} + d_\text{r}$.

\begin{figure}[]
\centering
\includegraphics[width=\linewidth]{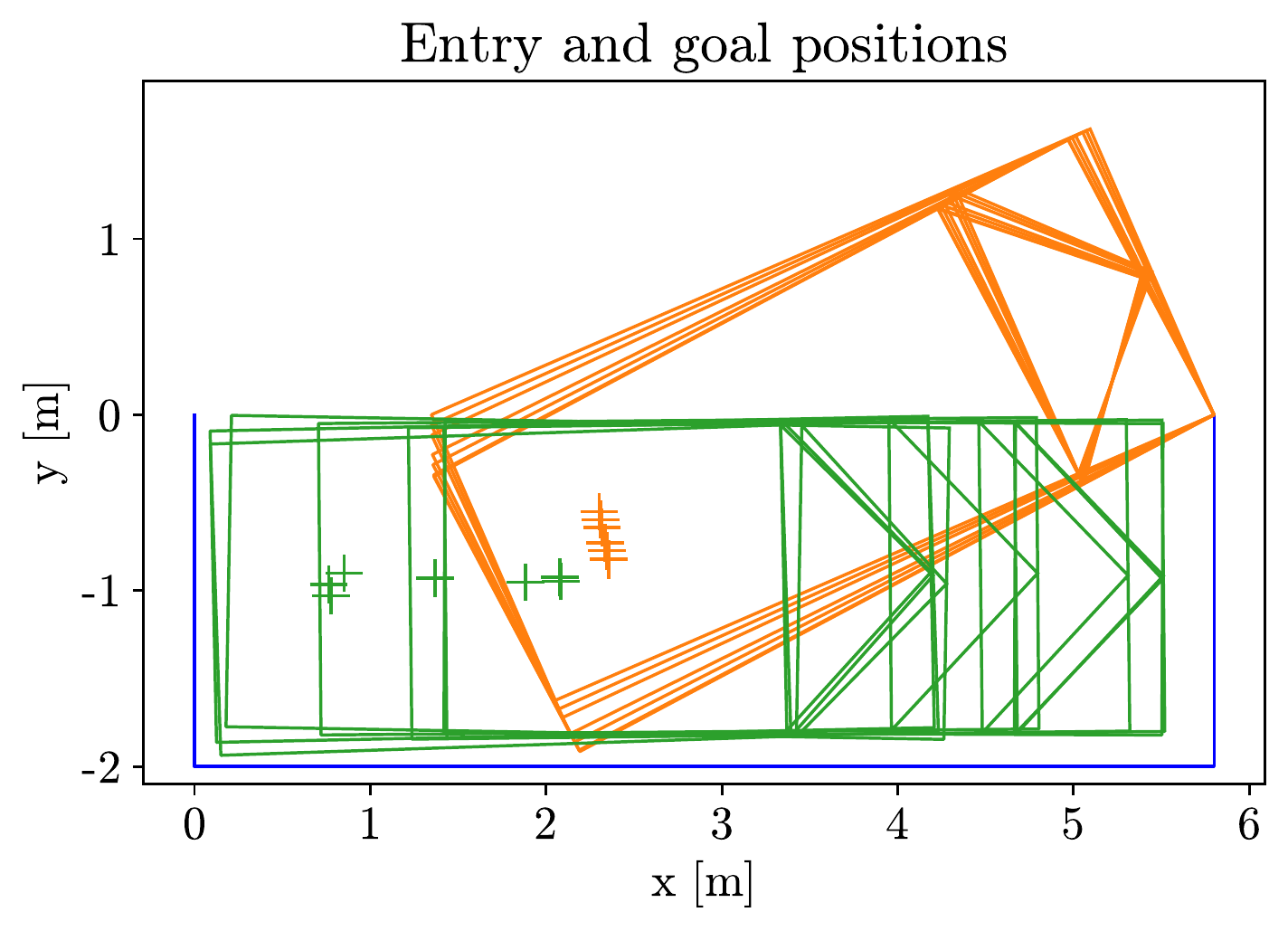}
\caption{Orange rectangles represent an example subset of computed entry positions, green rectangles are goal positions, and blue frame is parallel parking slot. The cross represents the middle of the rear axle.}
\label{f:ces}
\end{figure}

\emph{Goal position} is a car position $\cp$ for which $\mathcal{F}(\cp)$ is completely inside the parallel parking slot $P$. We can see an example of different green goal positions $\cp$ in \cref{f:ces}.

\begin{figure}[]
\centering
\includegraphics[width=\linewidth]{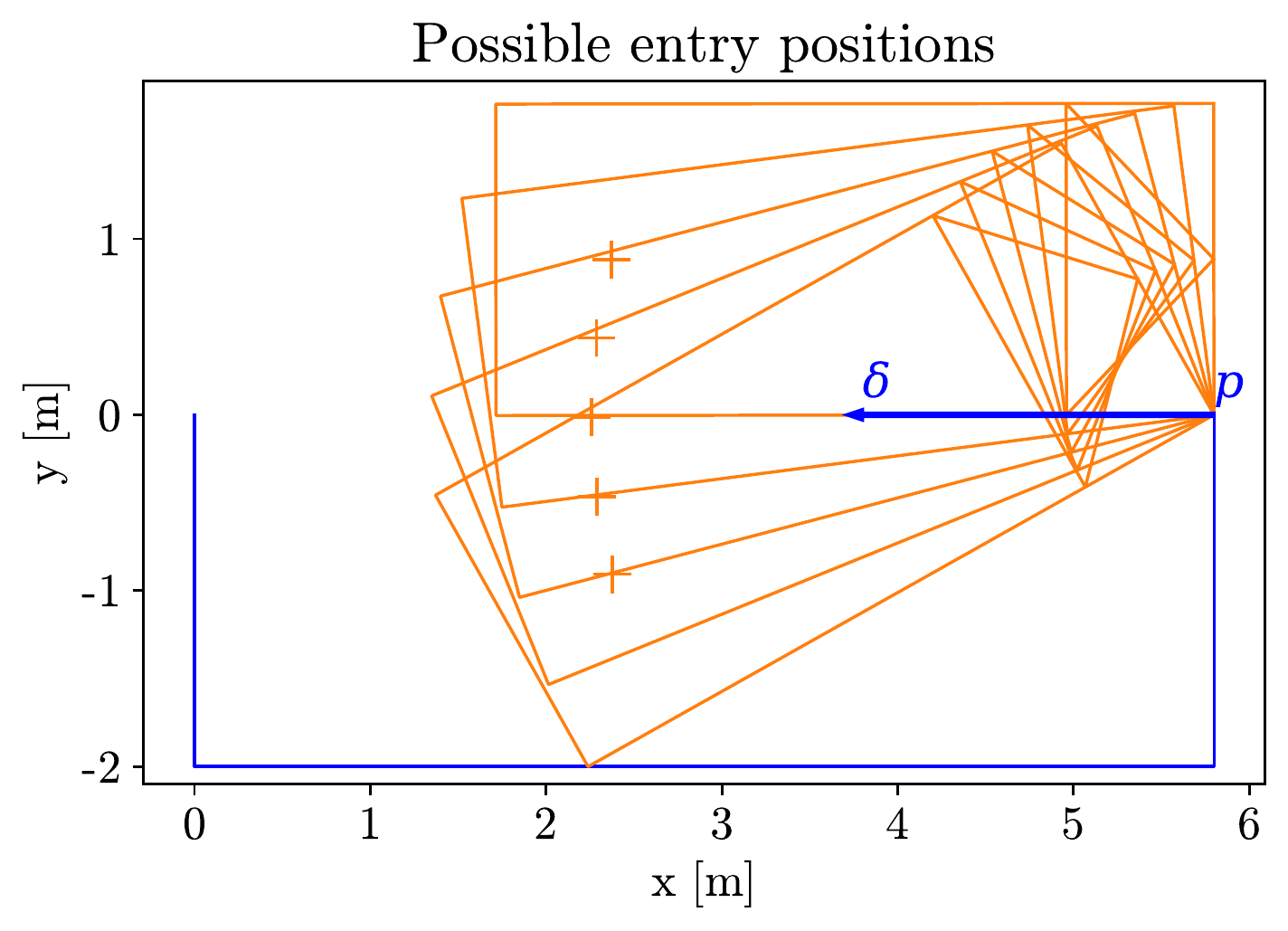}
\caption{Orange rectangles represent an example subset of possible entry positions, blue frame is parallel parking slot. The cross represents the middle of the rear axle.}
\label{f:ceps}
\end{figure}

\emph{Possible entry position} for parallel parking slot positioned on the right side of a road (for left-sided parking slot the definition would be symmetric) is a car position $\cep=(x,y,\theta, -1, +\phi_{\max})$ with the following properties: (i) right front corner of the car frame $\mathcal{F}(\cep)$ corresponds to the parking slot corner $p$, (ii) the car heading is in between the angle parallel to the entry side up to the angle perpendicular to the entry side, i.e., $\delta + \pi \leq \theta \leq \delta + 3/2\pi$, and (iii) car frame $\mathcal{F}(\ce^P)$ does not intersect with non-entry sides of the parking slot. We denote $\ceps$ a set of possible entry positions and we can see an example subset of $\ceps$ in \cref{f:ceps}.

\subsection{Problem statement}
\label{s:ps}

The main problem we solve is:
\begin{enumerate}
\item Given a parallel parking slot and car dimensions, find a set of entry positions $\ces\subset\ceps$ from which the car can park into the slot with the minimum number of direction changes.
\end{enumerate}

\noindent There are two additional related problems we solve along with the main one:

\begin{enumerate}
\setcounter{enumi}{1}
\item Given a parallel parking slot, car dimensions, and the maximum number of direction changes, what is a set of all entry positions $\ceas\subset\ceps$ from which the car can park into the slot?
\item Given the car dimensions and the maximum number of direction changes, what are the minimum dimensions (i.e., width and length) of a parallel parking slot the car can park into?
\end{enumerate}

\section{\uppercase{Parking Simulation Algorithms}}
\label{s:isp}
\label{s:fe}

In this section, we describe our parallel parking simulation algorithm, and we show how we use it to solve the problems from \cref{s:ps}. In \cref{s:ep}, we describe our algorithm for finding entry positions from which it is possible to park into the slot with the minimum number of direction changes, i.e., the main problem. In \cref{s:r}, we relax the condition on the minimum number of direction changes (problem 2), and in \cref{s:md} we describe how we use the algorithm to find the minimum dimensions of a parallel parking slot (problem 3).

\subsection{Finding entry positions}
\label{s:ep}

Our simulation algorithm first computes the set of possible entry positions $\ceps$. Then, the algorithm simulates backward-forward moves from each possible entry position $\cep\in\ceps$ until the car frame is completely inside the parking slot which mean the goal position is found. We can see the pseudocode in \cref{a:isp}. $\Delta_\theta$ is the increment to the car position heading for the consequent possible entry positions.

\begin{algorithm}[]
        \caption{Finding entry positions}
        \label{a:isp}
        \begin{itemize}
                \item \textbf{Input:} car dimensions $D$, parking slot $P$
                \item \textbf{Output:} entry positions $\ces$
        \end{itemize}
        \begin{algorithmic}[1]
          \State $\mathbf C_0\gets\{\cep(D, P, \theta)|\theta=\delta+\pi+m\cdot\Delta_\theta, m\in\mathbb{N}\}$
          \State $\mathbf H\gets\{(C_0, C_0, s, \phi, \Gamma)|C_0\in \mathbf C_0,\ s=-1,\ \phi=\phi_{max}, \Gamma=0\}$
          \State $\Gamma_{\max}\gets\infty$\label{a:isp:sg}
          \State $\mathbf I \gets \emptyset$ \Comment{Set of goal and entry positions}
          \While{$\mathbf I = \emptyset$}
          \For{$(C, C_0, s, \phi, \Gamma) \in \mathbf H$}
          \If{$\Gamma > \Gamma_{\text{max}}$}
            \State \textbf{continue}
          \EndIf
          \While{$F(C)$ not intersects with $P$}
          \State $C\gets f(C, s, \phi)$ \Comment{Simulate move}
          \EndWhile
          \If{$F(C)$ inside $P$}
          \State $\mathbf I\gets\mathbf I \cup (C, C_0)$
          \State $\Gamma_\text{max}\gets\Gamma$\label{a:isp:ug}
          \Else
          \State $\mathbf H\gets\mathbf H \cup (C, C_0, -s, -\phi, \Gamma+1)$
          \EndIf
          \EndFor
          \EndWhile
          \State \Return $\ces = \{C_0|(C,C_0)\in \mathbf I\}$
        \end{algorithmic}
\end{algorithm}

\subsection{Limit on the direction changes}
\label{s:r}

\cref {a:isp} finds the set of entry positions from which it is possible to park into the parking slot with the minimum number of direction changes. To find all entry positions $\ceas\subset\ceps$ for the given maximum number of direction changes, we provide $\Gamma_{\max}$ as the input to the \cref{a:isp} and remove $\Gamma_{\max}$ update in \cref{a:isp:ug}.

By removing $\Gamma_{\max}$ update, the computed set of entry positions does not need to be continuous, i.e., the difference in headings of neighboring entry positions could be greater than $\Delta_\theta$. We can see such a situation in \cref{f:ces}, where orange rectangles represent an example subset of computed entry positions where the continuity is broken approximately in the middle.

\subsection{Minimum slot dimensions}
\label{s:md}

To find the minimum dimensions (i.e., width and length) of a parallel parking slot for the given car dimensions and maximum number of direction changes $\Gamma_{\max}$, we run \cref{a:isp} for finding entry positions repeatedly, looping over different parking slot widths and lengths. First, we initialize the parking slot width $W=w$ and length $L=d_\text{f} + d_\text{r}$ to match the car dimensions. Then, we gradually increase the parking slot width and length by the width step $\Delta_\text{W}$ and the length step $\Delta_\text{L}$ respectively.

\section{\uppercase{Results}}
\label{s:eval}

In this section, we present the results of the computational experiments. In \cref{s:eval:c}, we compare our approach to the related works. In \cref{s:eval:r}, we show the range of all entry positions heading for the given maximum number of direction changes. Finally, we show the minimum dimensions of a parallel parking slot in \cref{s:eval:m}.

\begin{figure}
\centering
\includegraphics[width=\linewidth]{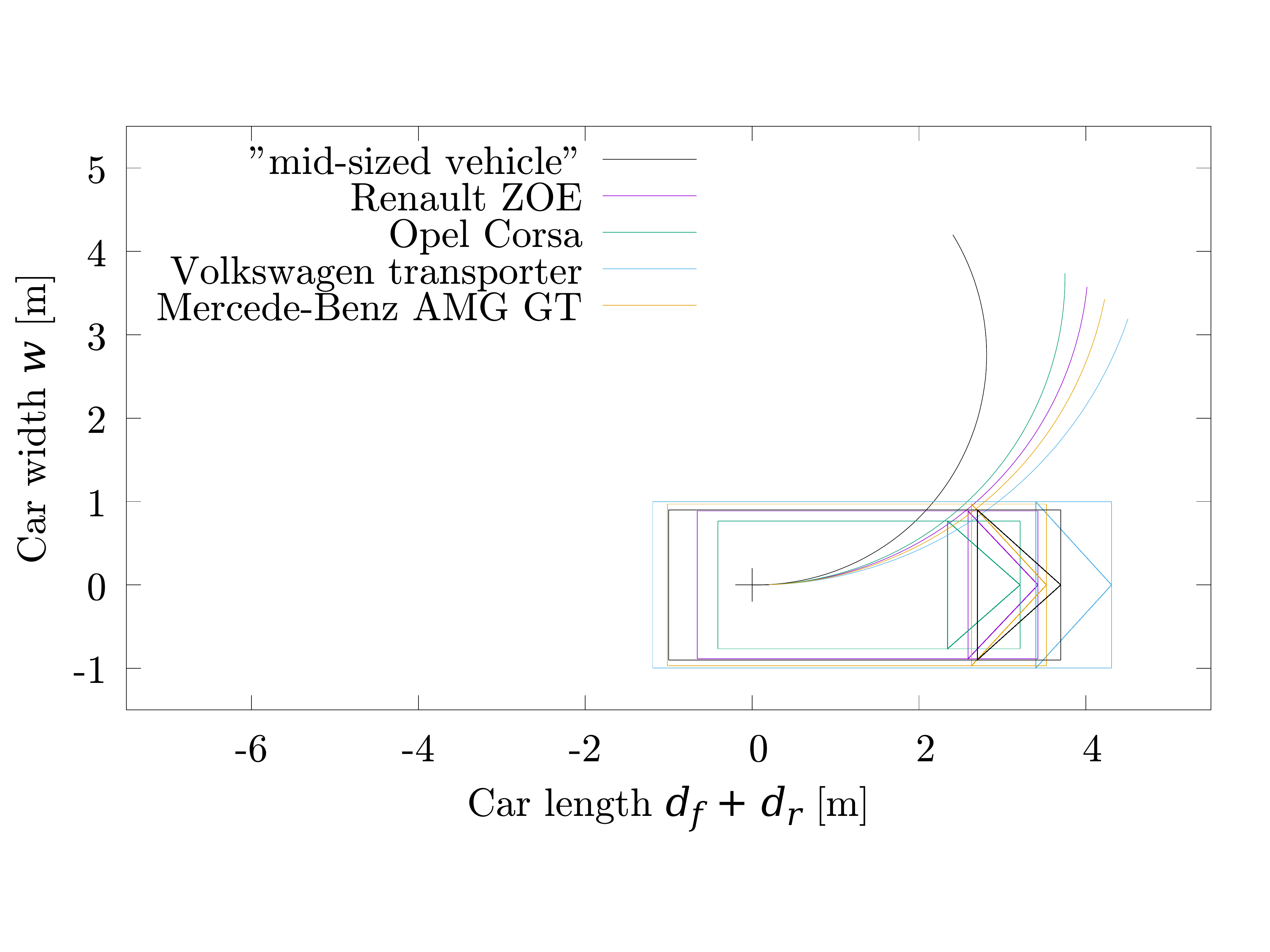}
\caption{Car frame and the simulated path given by the forward movement with the maximum steering angle for ``mid-sized vehicle'', Renault ZOE, Opel Corsa, Volkswagen transporter, and Mercedes-Benz AMG GT.}
\label{f:d}
\end{figure}

\begin{table*}[]
\caption{Car dimensions used in the computations.}
\centering
\begin{tabular}{|l|c|c|c|c|c|}
\hline
Car & $w$ [m] & $d_\text{f}$ [m] & $d_\text{r}$ [m] & $b$ [m] & $\phi_{\max}$ [$^{\circ}$] \\
\hline
\hline
``mid-sized vehicle'' & $1.8$ & $3.7$ & $1.0$ & $2.7$ & $45$ \\
Renault ZOE & $1.771$ & $3.427$ & $0.657$ & $2.588$ & $33$ \\
Opel Corsa & $1.532$ & $3.212$ & $0.410$ & $2.343$ & $32$ \\
Volkswagen transporter & $1.994$ & $4.308$ & $1.192$ & $3.400$ & $36$ \\
Mercedes-Benz AMG GT & $1.939$ & $3.528$ & $1.016$ & $2.630$ & $32$ \\
\hline
\end{tabular}
\label{t:d}
\end{table*}

We provide the results for five different cars. Their dimensions are shown in \cref{t:d}. In \cref{f:d}, we can see the car frames and the simulated paths given by the forward movement with the maximum steering angle. The width of the car is always without left and right rear view mirrors.

\subsection{Comparison to related works}
\label{s:eval:c}

\cite{zips_fast_2013} use the dimensions of the ``mid-sized vehicle'' and the parallel parking slot with the width $W=\SI{2.2}{m}$ and the length $L=\SI{5.1}{m}$ for their simulations. There is no change to the first phase of their algorithm in their follow-up work.

Along with the parallel parking slot and car dimensions, \cite{zips_fast_2013} are given the initial and goal positions. There is no further information or exact values provided in their simulation studies. They need at least $12$ direction changes to park in the slot.

We compute that it's possible to park into the slot they use with $10$ direction changes for the given car dimensions, so we conclude that the used goal position is not optimal.

\cite{vorobieva_automatic_2015} use the dimensions of the Renault ZOE for their experiments. They use \emph{several reversed trials} method to plan a path into the parallel parking slot.

We compare our implementation of \emph{several reversed trials} for goal positions used in \cite{vorobieva_automatic_2015} with goal positions computed by our approach.

The goal position used in \cite{vorobieva_automatic_2015} is not explicitly stated. They assume the goal position is known. For our computational experiments, we deduce the goal position from the context of the paper, i.e. the goal position for parallel parking slot positioned on the right side of a road has the following properties: (i) left side of the car frame given by the goal position corresponds to parking slot entry side, and (ii) rear side of the car frame given by the goal position corresponds to parking slot rear side.

The parallel parking slot has the width $W=\SI{2.0}{m}$, and we test the length of up to $L=\SI{5.8}{m}$ with the length step of $\Delta_\text{L}=\SI{0.01}{m}$.

\begin{figure}[]
\centering
\includegraphics[width=\linewidth]{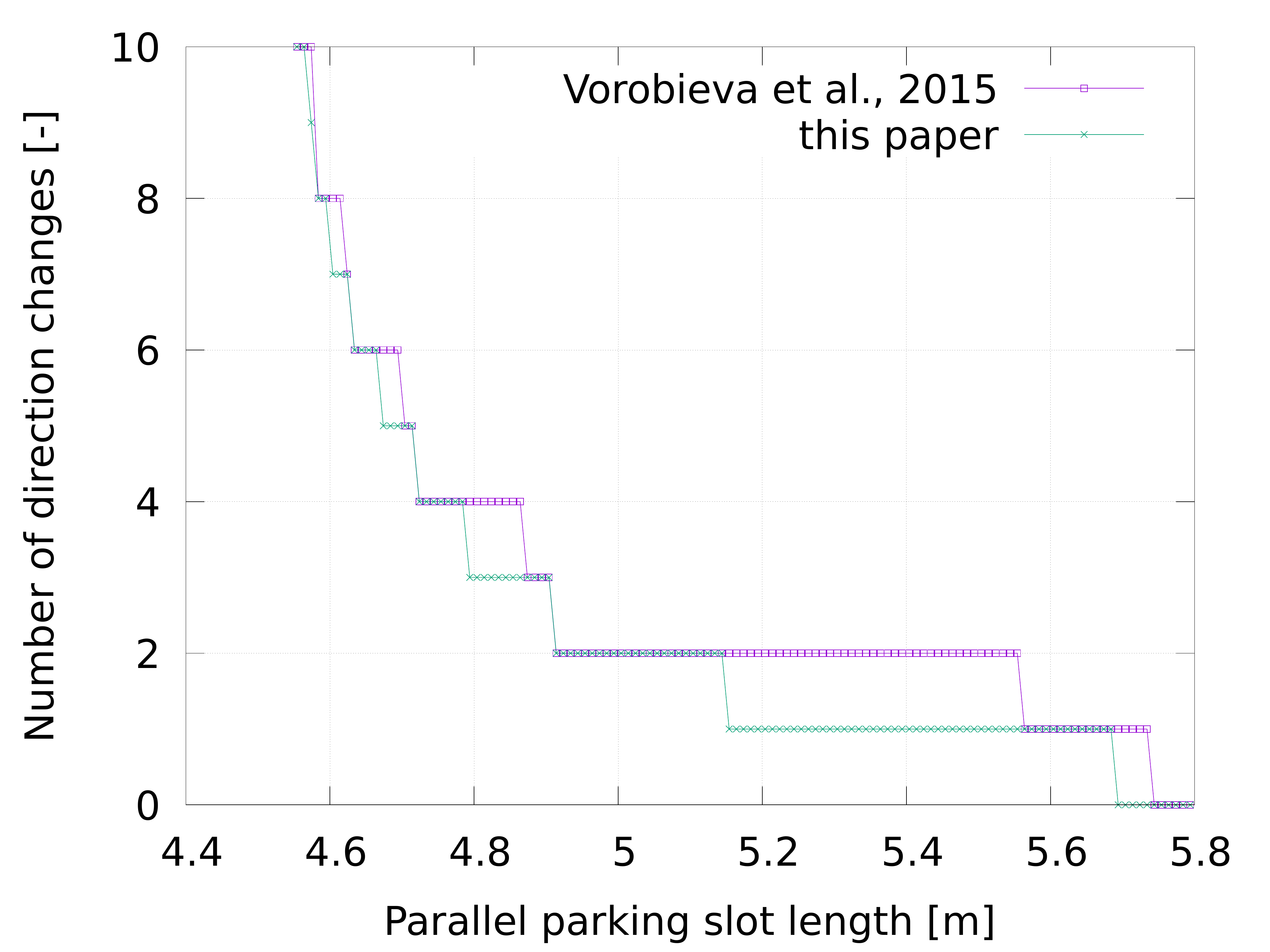}
\caption{Comparison of direction changes for Renault ZOE when \emph{several reversed trials} method is used with goal positions from \cite{vorobieva_automatic_2015} or with goal positions from this paper. Goal positions from \cite{vorobieva_automatic_2015} has the car frame aligned with the entry and rear sides of the parallel parking slot. The parallel parking slot has the width $W=\SI{2.0}{m}$ and variable length of up to $L=\SI{5.8}{m}$ with the length step of $\Delta_\text{L}=\SI{0.01}{m}$.}
\label{f:compare-to-vorobieva}
\end{figure}

In \cref{f:compare-to-vorobieva}, we can see that when goal position is given by our approach, it leads to less direction changes.

\cite{li_time-optimal_2016} use the dimensions of the Renault ZOE for their simulations. The parallel parking slot has the width $W=\SI{2.0}{m}$ and the length $L=\SI{4.5}{m}$. \cite{li_time-optimal_2016} need $19$ direction changes for the given parallel parking slot and car dimensions. Our approach results in $19$ direction changes for the given values, too.

\subsection{Limit on the direction changes}
\label{s:eval:r}

To compute the set of all entry positions for the given maximum number of direction changes, we use the car dimensions of the Renault ZOE.

\begin{figure}[]
\centering
\includegraphics[width=\linewidth]{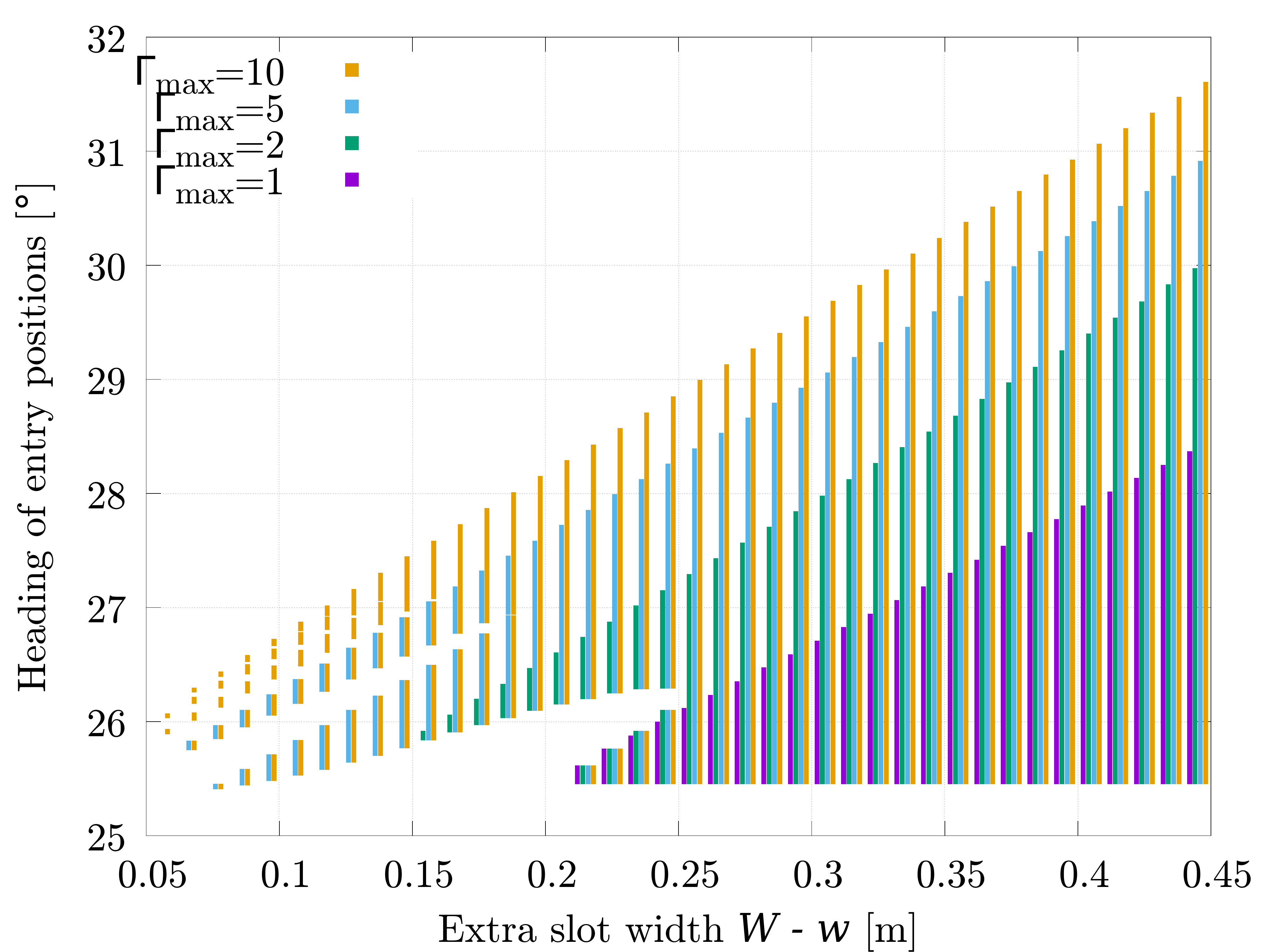}
\caption{Entry positions heading for Renault ZOE, the parallel parking slot with different extra slot width $W - w$ with the width step of $\Delta_\text{W}=\SI{0.01}{m}$, constant extra slot length $L - d_\text{f} - d_\text{r}=\SI{1.1}{m}$, and different maximum number of direction changes $\Gamma_{\max}$.}
\label{f:r}
\end{figure}

In \cref{f:r}, we can see the entry positions heading for the parallel parking slot with constant extra slot length $L - d_\text{f} - d_\text{r}=\SI{1.1}{m}$ and different maximum number of direction changes $\Gamma_{\max}=1$, $2$, $5$, and $10$ respectively.

We can see that the range of entry positions heading is wider for larger extra slot width $W - w$ and for increasing $\Gamma_{\max}$. Also, we can see that the entry positions are found for narrower slots for higher $\Gamma_{\max}$. Finally, the set of entry positions does not need to be continuous, i.e., the difference between two neighboring entry positions could be greater than $\Delta_\theta$. The range of entry positions heading then consists of the subranges.

\begin{figure}[]
\centering
\includegraphics[width=\linewidth]{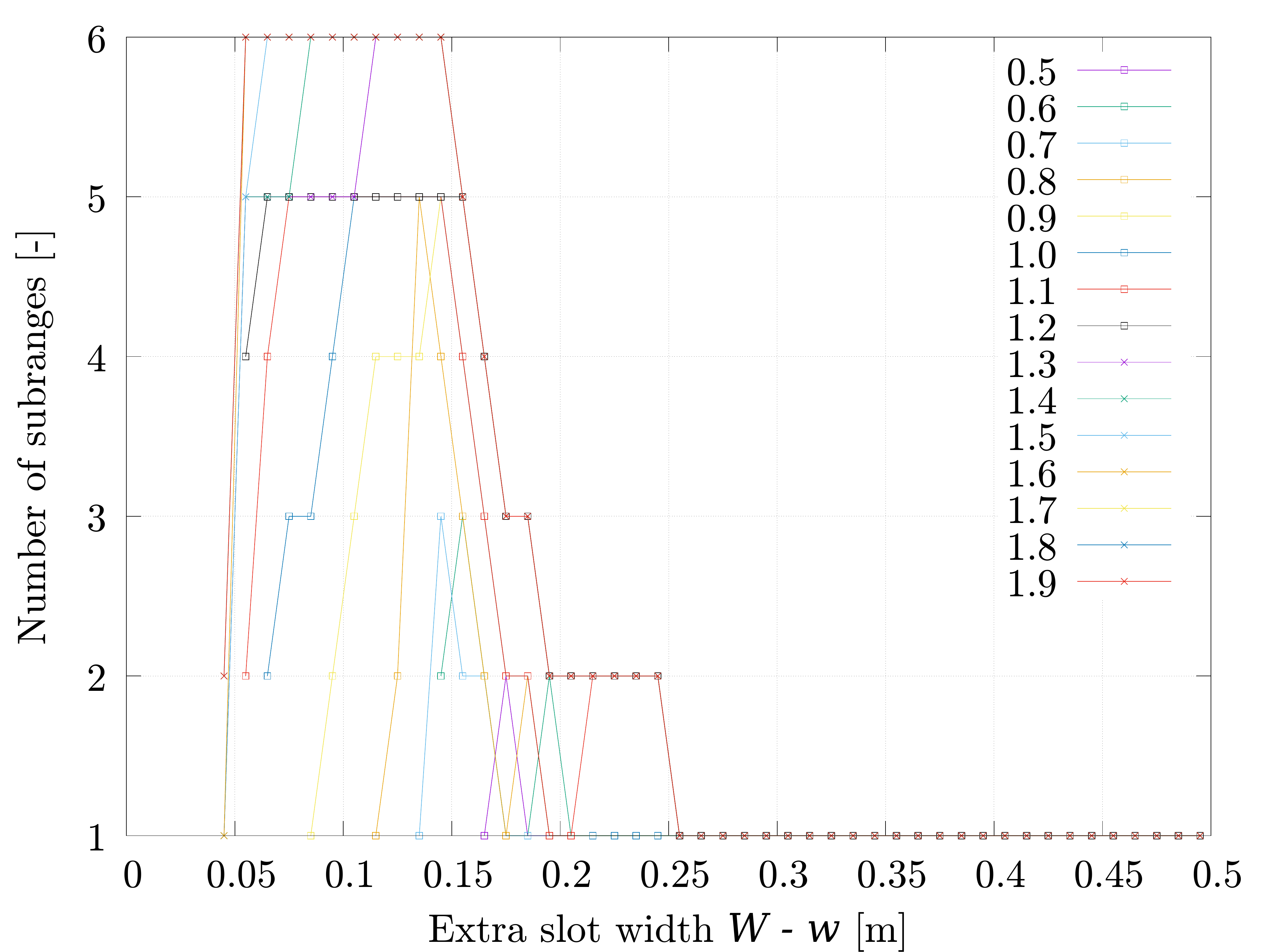}
\caption{The number of entry positions heading subranges for Renault ZOE, the parallel parking slot with different extra slot width $W - w$ with the width step of $\Delta_\text{W}=\SI{0.01}{m}$, extra slot length $L - d_\text{f} - d_\text{r}$ from $\SI{0.5}{m}$ up to $\SI{1.9}{m}$ with the length step of $\Delta_\text{L}=\SI{0.1}{m}$, and constant $\Gamma_{\max}=10$.}
\label{f:i}
\end{figure}

In \cref{f:i}, we can see the number of entry positions heading subranges for the parallel parking slot with different extra slot lengths $L - d_\text{f} - d_\text{r}$ from $\SI{0.5}{m}$ up to $\SI{1.9}{m}$ and constant $\Gamma_{\max}=10$.

We can see that there is only one range of entry positions heading for wide enough slots. The number of subranges increases for narrower slots but decreases again for the slots that are too narrow.

\subsection{Minimum slot dimensions}
\label{s:eval:m}

To compute the minimum dimensions of a parallel parking slot, we use the possible entry position heading step $\Delta_\theta=\SI{e-4}{rad}$, parking slot width step $\Delta_\text{W}=\SI{0.01}{m}$, and the length step $\Delta_\text{L}=\SI{0.01}{m}$.

\begin{figure}[]
\centering
\includegraphics[width=\linewidth]{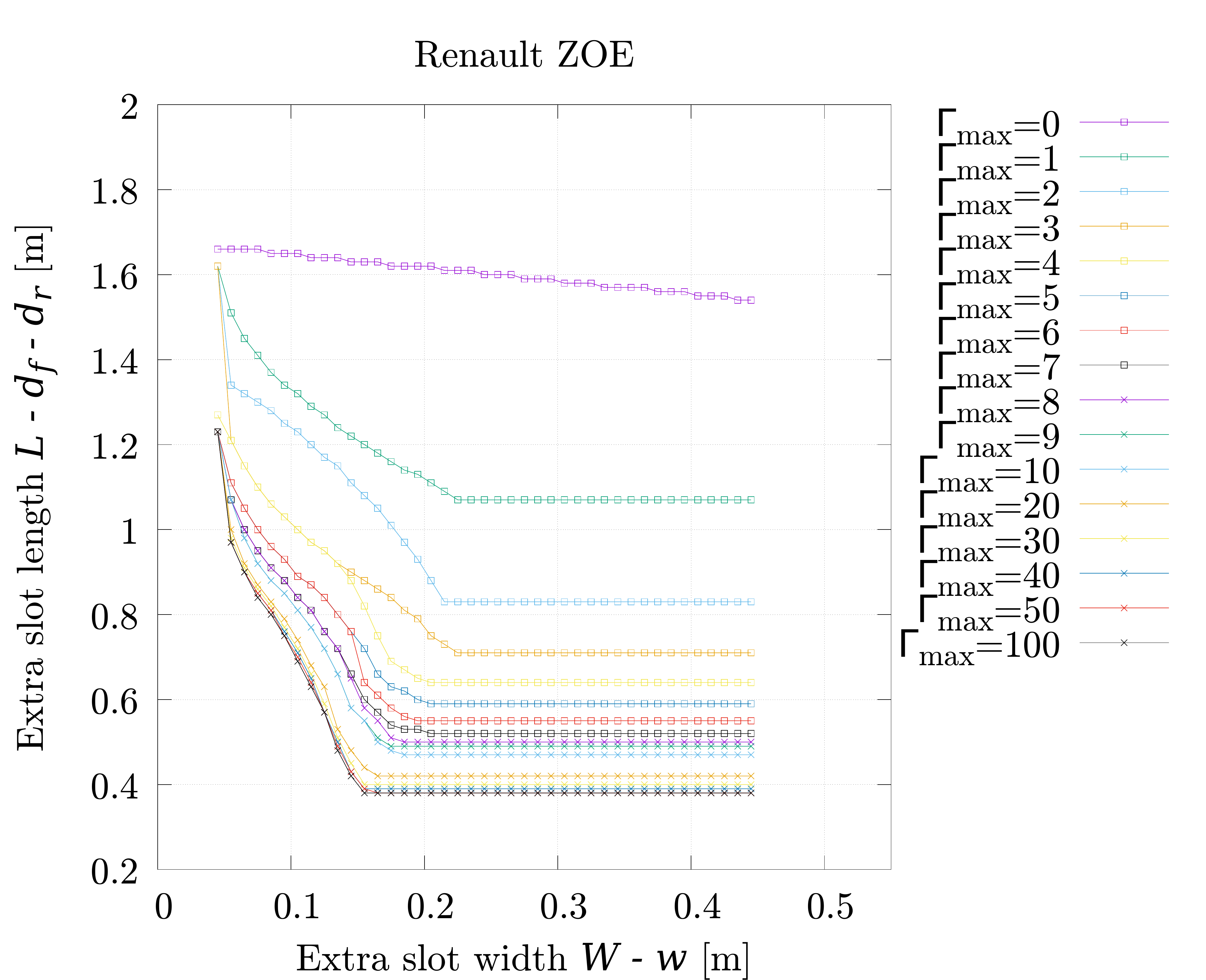}
\caption{Computed minimum dimensions of a parallel parking slot for Renault ZOE and different maximum number of direction changes. The first line at the top of the graph shows the minimum dimensions when no direction change is allowed ($\Gamma_{\max}=0$), followed by the maximum number ($\Gamma_{\max}$) of 1, 2, 3, 4, 5, 6, 7, 8, 9, 10, 20, 30, 40, 50, and 100 direction changes. The width step $\Delta_\text{W}=\SI{0.01}{m}$ and the length step $\Delta_\text{L}=\SI{0.01}{m}$.}
\label{f:v-pp}
\end{figure}

In \cref{f:v-pp}, we can see the computed minimum dimensions of a parallel parking slot for Renault ZOE and different maximum number of direction changes $\Gamma_{\max}$.

Renault ZOE needs to perform at least $\Gamma=30$ direction changes when parking into the parallel parking slot with extra slot length $L - d_\text{f} - d_\text{r}=\SI{0.4}{m}$. Also, there is a negligible improvement of $\SI{0.02}{m}$ when increasing the number of direction changes to $\Gamma=100$.

\begin{figure}[]
\centering
\begin{tabular}{c}
\includegraphics[width=\linewidth]{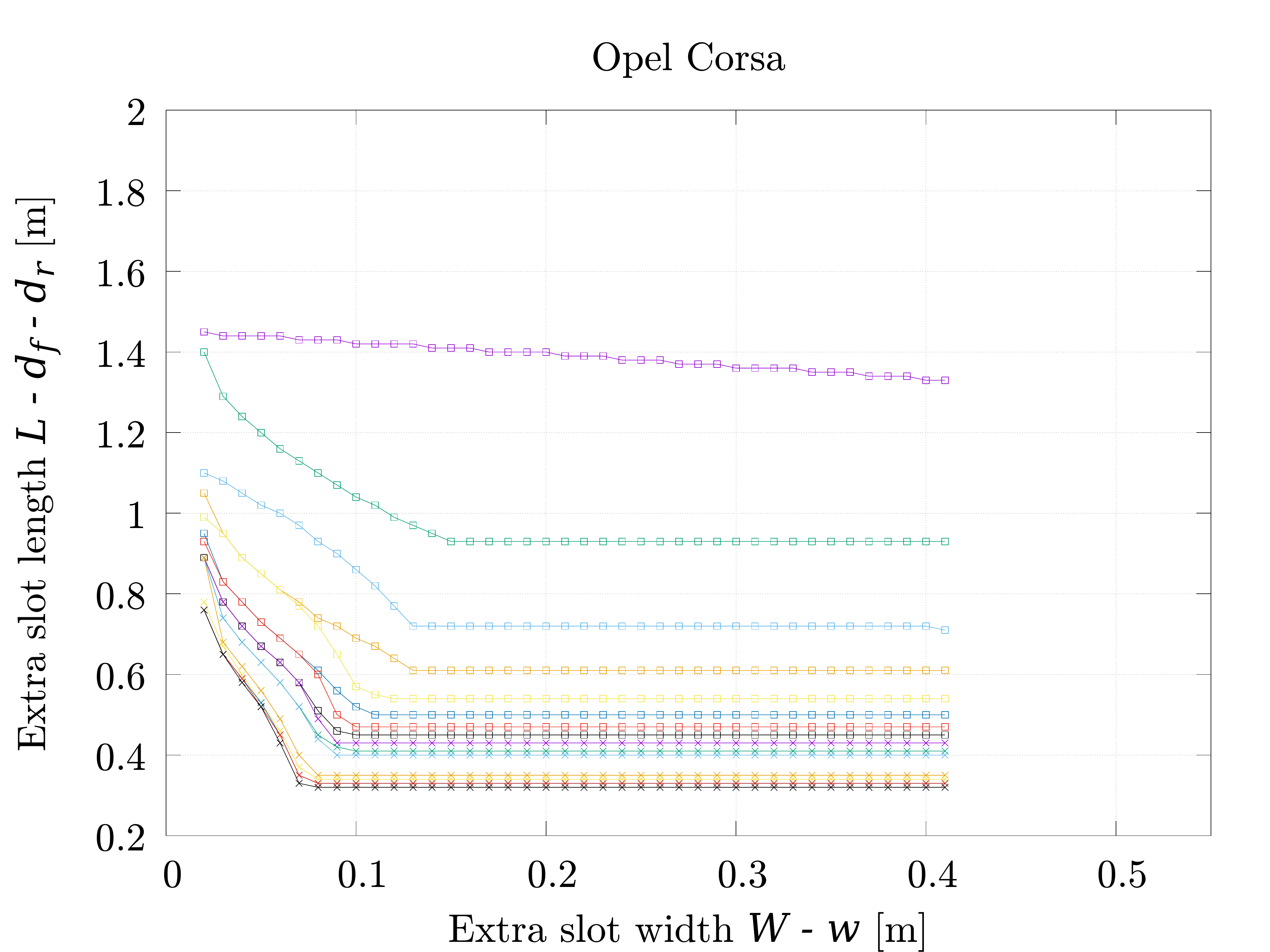} \\
\includegraphics[width=\linewidth]{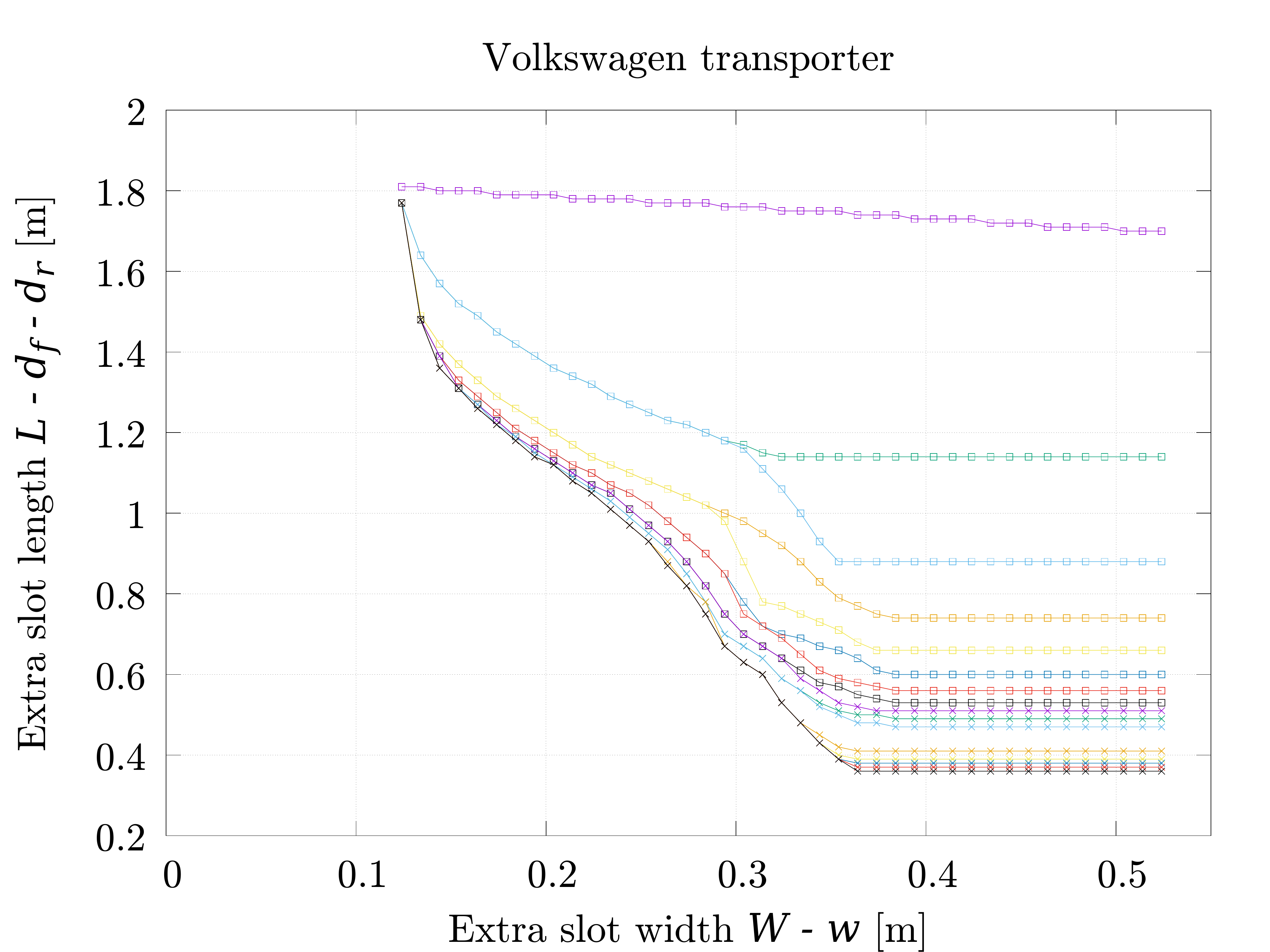} \\
\includegraphics[width=\linewidth]{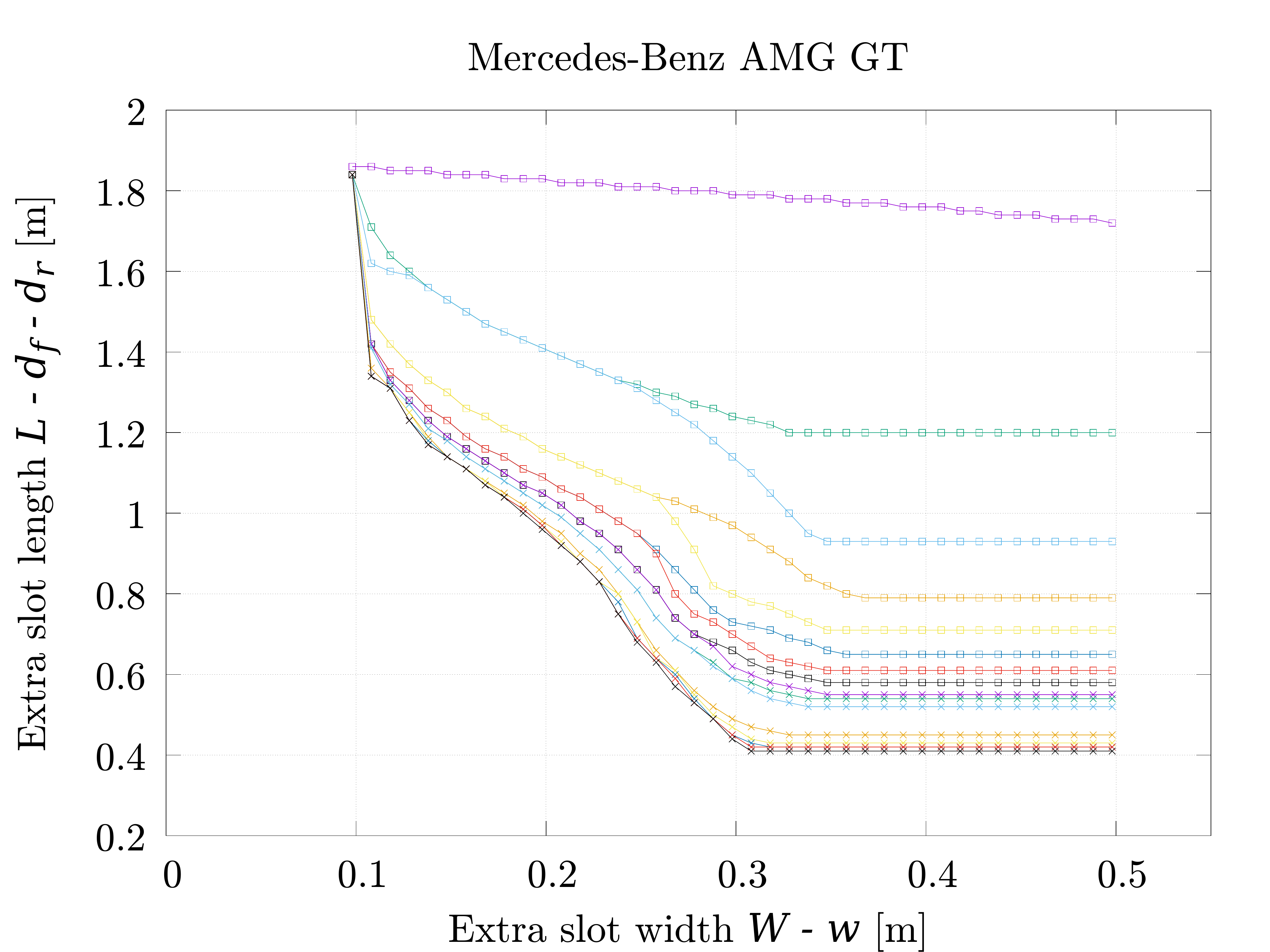}
\end{tabular}
\caption{Computed minimum dimensions of a parallel parking slot for Opel Corsa, Volkswagen transporter, Mercedes-Benz AMG GT, and different maximum number of direction changes.}
\label{f:oth}
\end{figure}

In \cref{f:oth}, we can see the minimum dimensions of a parallel parking slot for Opel Corsa, Volkswagen transporter, and Mercedes-Benz AMG GT.

Opel Corsa can park into the parallel parking slot with extra slot length of $\SI{0.4}{m}$ with $\Gamma=10$ direction changes. Volkswagen transporter needs the same number of direction changes $\Gamma=30$ as the Renault ZOE, although it requires more extra slot width. It is not possible to park Mercedes-Benz AMG GT into the parallel parking slot with extra slot length of $\SI{0.4}{m}$ within the reasonable number of backward-forward direction changes.

The improvement to extra slot length for $\Gamma>20$ is negligible, i.e., in the order of $\SI{e-2}{m}$. The shape of the slot's minimum dimensions curve depends on whether the number of backward-forward direction changes is even or odd.

\begin{figure}[]
\centering
\includegraphics[width=\linewidth]{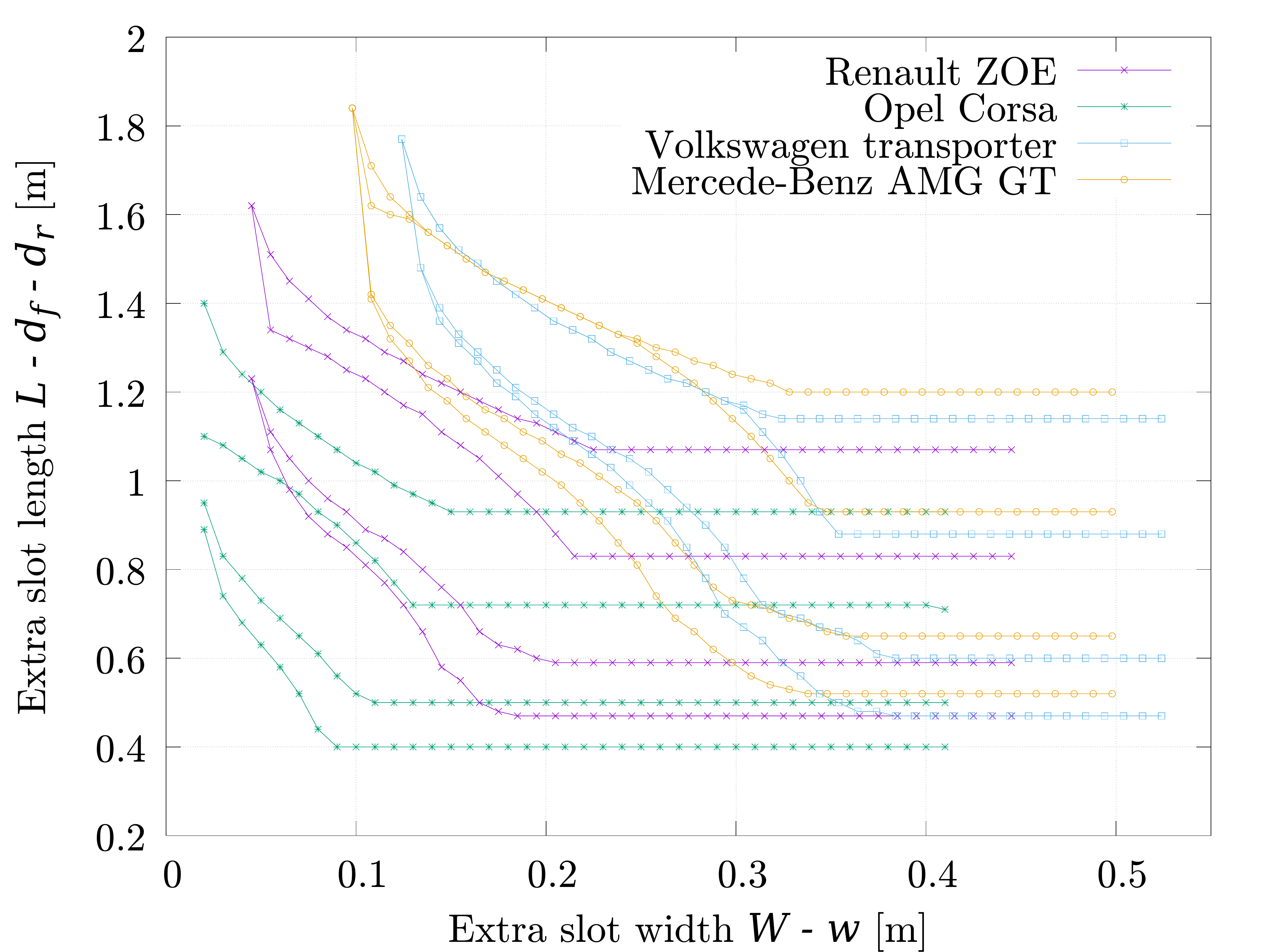}
\caption{Computed minimum dimensions of a parallel parking slot for Renault ZOE, Opel Corsa, Volkswagen transporter, and Mercedes-Benz AMG GT. The width step $\Delta_\text{W}=\SI{0.01}{m}$ and the length step $\Delta_\text{L}=\SI{0.01}{m}$. The maximum number of direction changes is $\Gamma_{\max}=1$, $2$, $5$, and $10$ respectively.}
\label{f:a-pp}
\end{figure}

\cref{f:a-pp} shows the minimum dimensions of a parallel parking slot for all the cars. The maximum number of direction changes shown in the figure is $\Gamma_{\max}=1$, $2$, $5$, and $10$ respectively.

\section{\uppercase{Conclusion}}
\label{s:concl}

In this paper, we re-formulate the path planning problem for parallel parking: Given a parallel parking slot and car dimensions, we compute a set of entry positions to the parking slot from which it is possible to park into the slot with the minimum number of backward-forward direction changes. We also use our approach to compute the minimum dimensions (i.e., width and length) of a parallel parking slot for the given car dimensions.

Our algorithm is designed for offline use. By precomputing entry positions for different parallel parking slots and car dimensions and using them in online planning, we can speed up the path planning as it does not need to plan the path inside the parking slot. Parking assistants using our results can ensure that when the automated vehicle reaches one of the entry positions, it can optimally park into the parallel parking slot.

The range of entry positions from which the given car can park optimally depends on the slot dimensions. The range is wider for larger slots and a higher number of backward-forward direction changes as shown in \cref{s:eval}.

By precomputing minimum dimensions of different parallel parking slots, we can simplify the decision of whether it is possible to park into the slot.

Using more than $20$ direction changes does not lead to significant reduction of the parking slot size. The shape of the slot's minimum dimensions curve depends on whether the number of backward-forward direction changes is even or odd.

\vfill
\section*{\uppercase{Acknowledgements}}

Research leading to these results has received funding from the EU ECSEL Joint Undertaking and the Ministry of Education of the Czech Republic under grant agreement 826452, 8A19011, and MSMT-24623/2019-4/11 (project Arrowhead Tools).

\bibliographystyle{apalike}
{\small
\bibliography{ms}}

\section*{\uppercase{Appendix}}

In this section, there are more figures we used for the presentation.

\onecolumn

\begin{figure}
\centering
\includegraphics[
  width=\textwidth,
  keepaspectratio
]{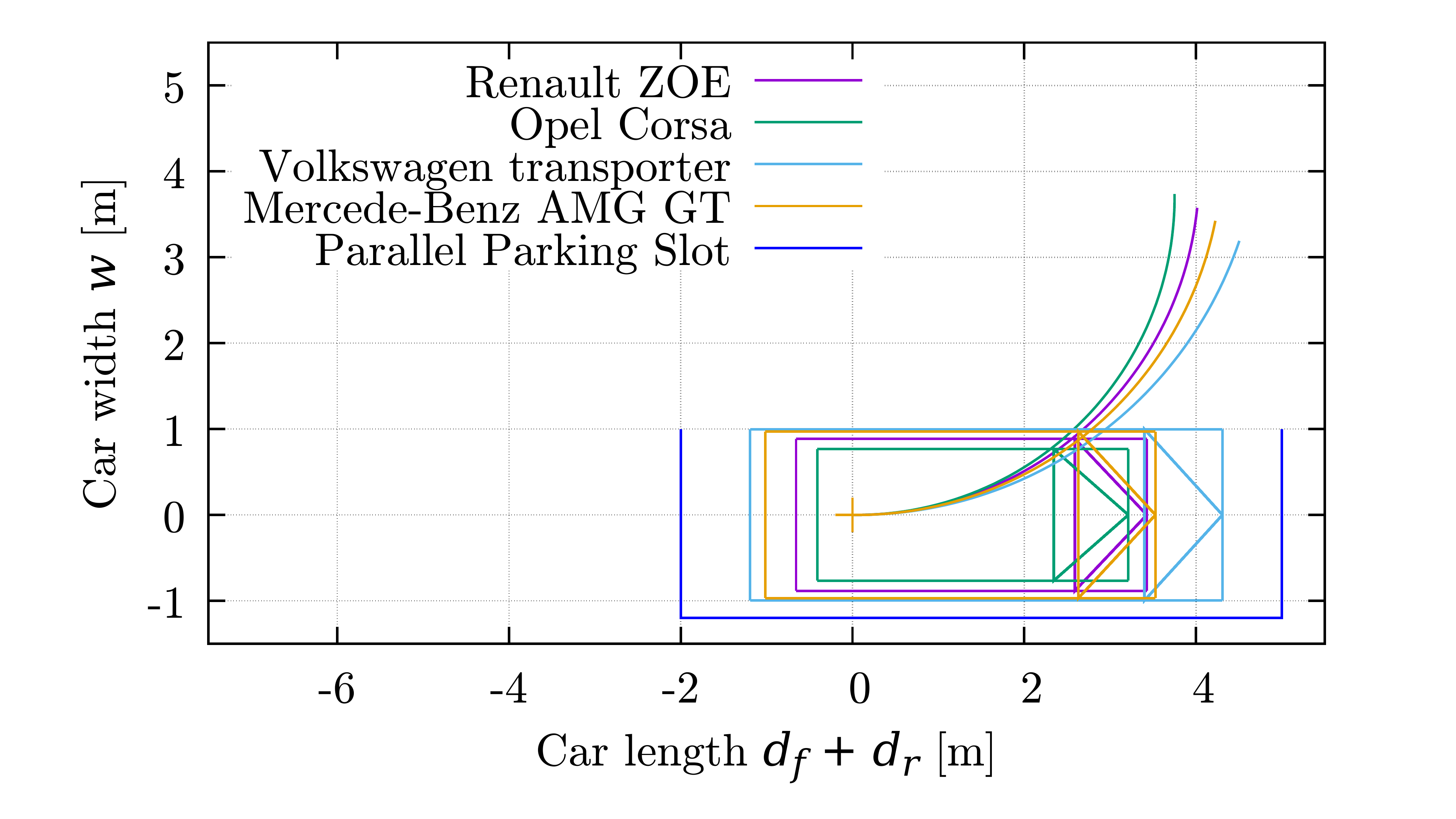}
\caption{Car frame and the simulated path given by the forward movement with the maximum steering angle for Renault ZOE, Opel Corsa, Volkswagen transporter, and Mercedes-Benz AMG GT.}
\end{figure}

\begin{figure}
\centering
\includegraphics[
  width=\textwidth,
  keepaspectratio
]{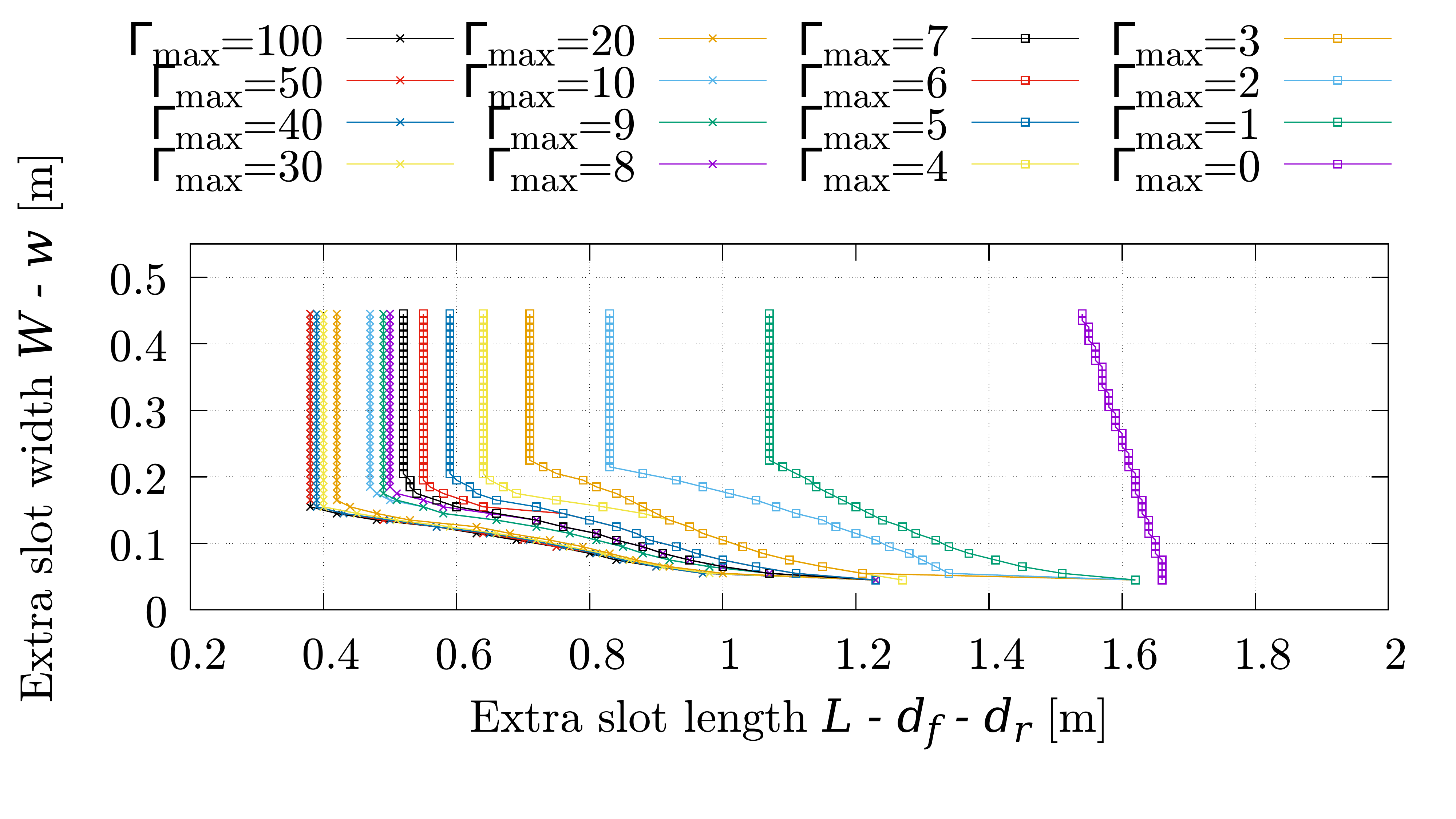}
\caption{Computed minimum dimensions of a parallel parking slot for Renault ZOE and different maximum number of direction changes. The first line at the top of the graph shows the minimum dimensions when no direction change is allowed ($\Gamma_{\max}=0$), followed by the maximum number ($\Gamma_{\max}$) of 1, 2, 3, 4, 5, 6, 7, 8, 9, 10, 20, 30, 40, 50, and 100 direction changes. The width step $\Delta_\text{W}=\SI{0.01}{m}$ and the length step $\Delta_\text{L}=\SI{0.01}{m}$.
\\\\
We can see that parking in the slot with \SI{40}{cm} extra slot length is hard, that there is no significant improvement for more than \SI{20}{} backward-forward direction changes, and that there is no improvement for more than \SI{25}{cm} extra slot width.}
\end{figure}

\begin{figure}
\centering
\includegraphics[
  width=\textwidth,
  keepaspectratio
]{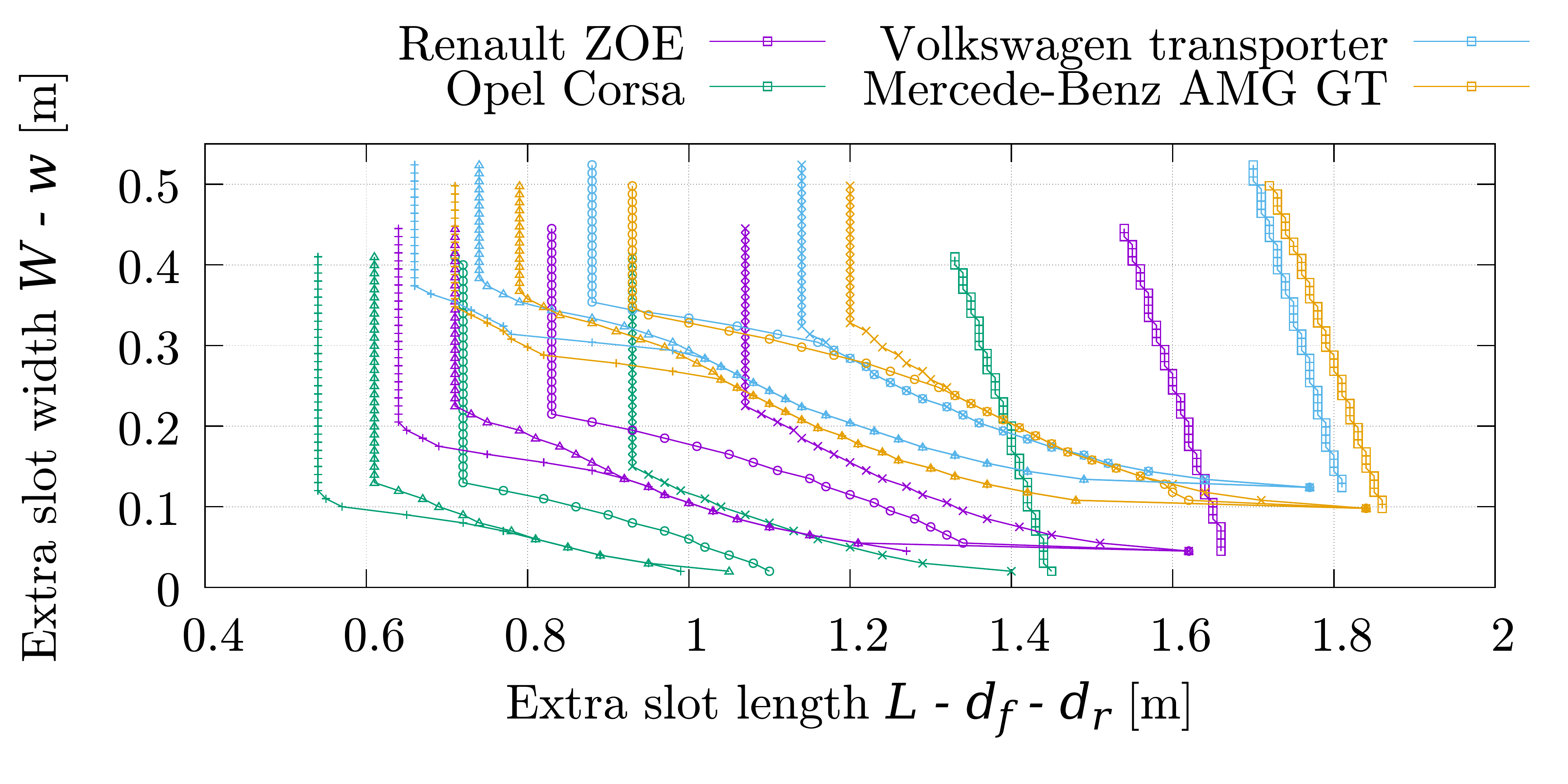}
\caption{Computed minimum dimensions of a parallel parking slot for Renault ZOE, Opel Corsa, Volkswagen transporter, and Mercedes-Benz AMG GT. The width step $\Delta_\text{W}=\SI{0.01}{m}$ and the length step $\Delta_\text{L}=\SI{0.01}{m}$. The maximum number of direction changes $\Gamma_{\max}$ is $0=\square$, $1=\times$, $2=\circ$, $3=\triangle$, and $4=+$.
\\\\
We can see that extra slot length is more significant than extra slot width. Interesting thing is that parking of the transporter requires less extra slot length than parking of the sport car.}
\end{figure}

\begin{figure}
\centering
\includegraphics[
  width=\textwidth,
  keepaspectratio
]{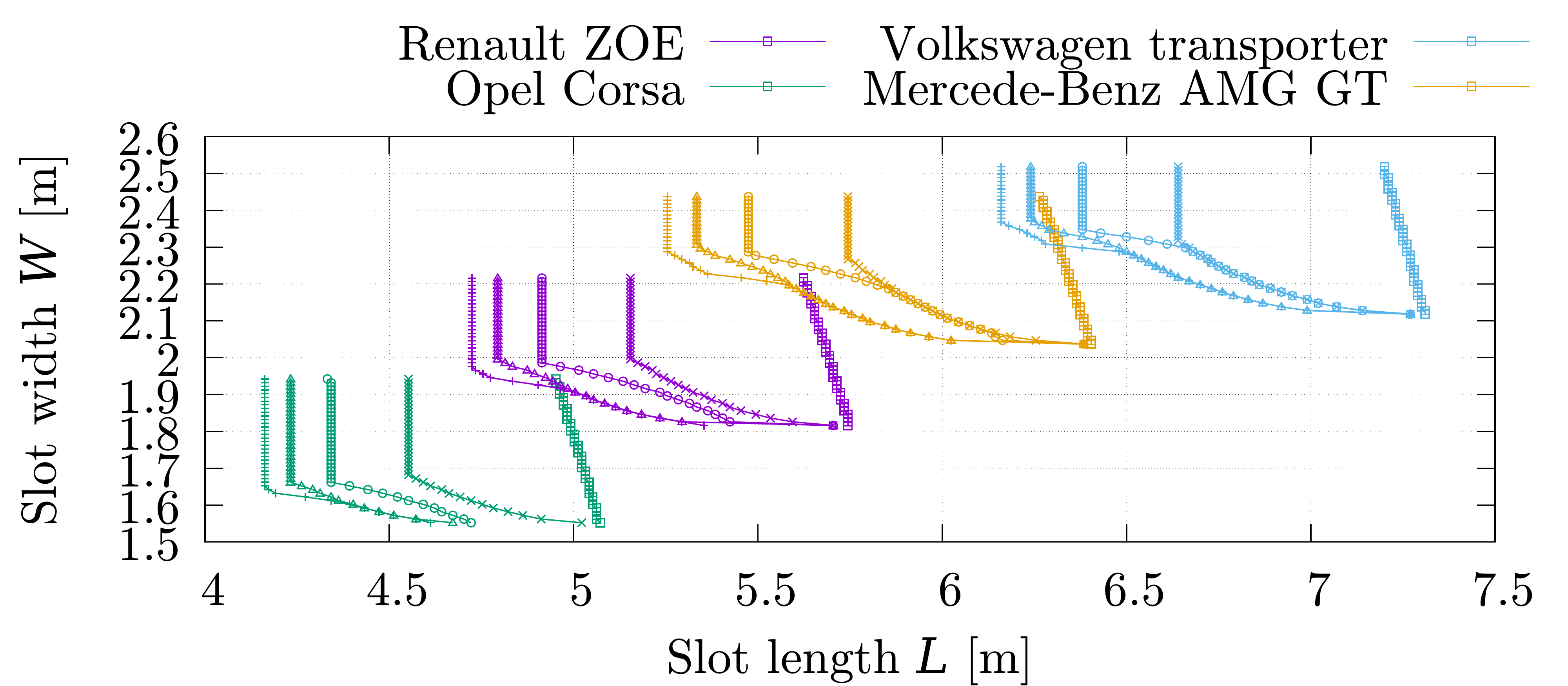}
\caption{Computed absolute dimensions of minimum parallel parking slots for Renault ZOE, Opel Corsa, Volkswagen transporter, and Mercedes-Benz AMG GT. The width step $\Delta_\text{W}=\SI{0.01}{m}$ and the length step $\Delta_\text{L}=\SI{0.01}{m}$. The maximum number of direction changes $\Gamma_{\max}$ is $0=\square$, $1=\times$, $2=\circ$, $3=\triangle$, and $4=+$.}
\end{figure}

%

\end{document}